\documentclass{isprs} 
\usepackage{setspace}
\usepackage{geometry} 
\usepackage{epstopdf}
\usepackage[labelsep=period]{caption}  
\usepackage[british]{babel} 
\usepackage[hang]{footmisc}

\usepackage{lipsum}

\usepackage{graphicx}
\graphicspath{{./images/}}
\DeclareGraphicsExtensions{.pdf,.png,.jpg}
\usepackage{subcaption}

\usepackage[hidelinks]{hyperref}
\hypersetup{
  colorlinks   = true, 
  urlcolor     = blue,
  linkcolor    = blue, 
  citecolor    = blue 
}

\usepackage{bm}
\usepackage{amsmath}

\usepackage{booktabs}
\usepackage{multirow}

\begin{document}

\title{Fast and Regularized Reconstruction of Building Fa\c{c}ades from Street-View Images using Binary Integer Programming}

\author{
    Han Hu\textsuperscript{a, b, }\thanks{Corresponding Author} , Libin Wang\textsuperscript{b}, Mier Zhang\textsuperscript{b}, Yulin Ding\textsuperscript{a, b}, Qing Zhu\textsuperscript{b}
}

\address{
	\textsuperscript{a }State Key Laboratory of Rail Transit Engineering
	Informatization, China Railway First Survey and Design\\ Institute Co. Ltd., Xi’an, China \\
    \textsuperscript{b }Faculty of Geosciences and Environmental Engineering, Southwest Jiaotong University, Chengdu\\
    (han.hu@swjtu.edu.cn, wlb@my.swjtu.edu.cn, mierzhang@my.swjtu.edu.cn, rainforests@126.com, zhuq66@263.net)
}


\commission{II, }{II} 
\workinggroup{II/4} 
\icwg{}   

\abstract{
    Regularized arrangement of primitives on building fa\c{c}ades to aligned locations and consistent sizes is important towards structured reconstruction of urban environment. Mixed integer linear programing was used to solve the problem, however, it is extreamly time consuming even for state-of-the-art commercial solvers. Aiming to alleviate this issue, we cast the problem into binary integer programming, which omits the requirements for real value parameters and is more efficient to be solved . Firstly, the bounding boxes of the primitives are detected using the YOLOv3 architecture in real-time. Secondly, the coordinates of the upper left corners and the sizes of the bounding boxes are automatically clustered in a binary integer programming optimization, which jointly considers the geometric fitness, regularity and additional constraints; this step does not require \emph{a priori} knowledge, such as the number of clusters or pre-defined grammars. Finally, the regularized bounding boxes can be directly used to guide the fa\c{c}ade reconstruction in an interactive envinronment. Experimental evaluations have revealed that the accuracies for the extraction of primitives are above 0.82, which is sufficient for the following 3D reconstruction. The proposed approach only takes about $ 10\% $ to $ 20\% $ of the runtime than previous approach and reduces the diversity of the bounding boxes to about $20\%$ to $50\%$.
}


\maketitle

\section{Introduction}
\label{s:intro}

\sloppy

Reconstruction of building fa\c{c}ades is one of the key steps towards complete reconstruction of a LOD-3 (Level-of-Details) model in CityGML protocol \cite{groger2012citygml}. Semantic objects such as windows, doors, and balconies are important components of a building fa\c{c}ade. Extracting them \cite{hoegner2015building} and arranging them in a regularized manner \cite{hensel2019facade} are two important steps towards structured LOD-3 reconstruction \cite{zhu2020interactive}. And the street-view image is arguably the best option for the above objectives due to the public availability and effectiveness in collecting, such as the Google street map \cite{anguelov2010google}.

For the detection of semantic objects in street-view images, classical methods include the use of projected histograms \cite{lee2004extraction,kostelijk2012semantic}, gradient projection, K-means clustering \cite{recky2010windows}, correlation coefficient \cite{mayer2007building}, perceptual grouping \cite{sirmacek2011detection} and \emph{etc.}. Such methods do not consider the structural and spatial distribution of the semantic objects. Recently, methods based on deep learning \cite{mathias2016atlas,liu2017deepfacade} have been widely used to extract the semantic objects on building fa\c{c}ade, which have achieved impressive results on images with projective distortion and scale difference; but the regularities of semantic objects have not been considered yet. 

In general, these semantic objects should conform to certain regularities, such as aligned locations and consistent sizes. However, due to the characteristics of projection distortion and complex background, the geometric attributes of the extracted primitives in images of buildings fa\c{c}ade are generally deviated slightly from the expected. Although the regularization of 2D boundaries, such as edges of buildings, are widely studied in the community \cite{xie2018hierarchical}, the approaches cannot be directly adopted. In addition, the regular arrangements of fa\c{c}ades can also be learned for specific scenarios \cite{dehbi2011learning,dehbi2017statistical}; however, the learned models can only be used in inductive fashion, \emph{e.g.} it does not generalize to unseen data. 

Recently, a general and promising approach to align different objects of building fa\c{c}ades using Mixed Integer Linear Programming (MILP) was proposed \cite{hensel2019facade}. However, in our practice the MILP is too complex to solve, which requires prohibitively high runtime consumption. Because we are aiming to integrate the pipeline into an interactive reconstruction environment, at least near real-time response of the solver is required. To solve this issue, we reformulate the problem as a Binary Integer Programming (BIP), with all the unknowns in the binary space of $ \{0,~1\} $, and the objective can be expressed explicitly as logical operations of the binary variables. Rather than MILP, the BIP is relatively more efficient to be handled by state-of-the-art optimization routines \cite{gleixner2018scip,gurobi2014gurobi}.

In summary, this paper proposes a fast and regularized reconstruction methods for the fa\c{c}ades of buildings from street-view images. Firstly, we extract typical fa\c{c}ade primitives using real-time object detection pipeline, \emph{e.g.} the YOLOv3 architectural \cite{redmon2016you,redmon2018yolov3}. Secondly, the positions and sizes of the primitives are clustered using BIP by optimizing two competing desires of retaining the best fitness and regularities, for which we require no extra information of the fa\c{c}ades. At last, the primitives after clustering are reconstructed in an interactive environment, \emph{e.g.} SketchUp, by substituting each clustered primitive with a pre-built component model or interactively sketching the component on street-view images. 

\section{Related Works}
\label{s:relat}

A lot of works have been devoted to extraction and segmentation of building fa\c{c}ades, in the communities of photogrammetry, computer vision and computer graphics. With regard to detecting fa\c{c}ade objects from images, in recent years, various deep learning architectures, such as CNN \cite{krizhevsky2012imagenet} and RNN \cite{graves2008novel}, have achieved impressive results for various computer vision tasks, such as image classification \cite{chan2015pcanet} and object detection \cite{girshick2015region}. Although earlier CNN architectures can greatly improve the accuracy of object detection, the detection rate is very slow. This is because that several segregated steps \cite{girshick2015region} are used, including generation of proposals and classification of the regions. For this reason, the usage in applications requiring real-time responses is limited. The YOLO (You Only Look Once) network \cite{redmon2016you,redmon2018yolov3}, as the name suggested, only requires a single integrated forward passing in the testing stage and achieves real-time detection rates for off-the-shelf video sensors. The incrementally upgraded YOLOv3 \cite{redmon2018yolov3}, due to the integration of ResNet \cite{he2016deep}, FPN (Feature Pyramid Network) \cite{lin2017feature}, and binary cross entropy loss, greatly improves both detection speed and detection accuracy. In the meantime, it has also increased the performance on small targets, which is suitable for detecting semantic objects with complex repeating structures on the building fa\c{c}ade. And therefore, this paper adopts the YOLOv3 as the backbone for the detection of the primitives.

With regard to the regular arrangements of objects, based on explicit or implicit procedural methods, the structure of fa\c{c}ade was inferred through grammatical rules, including random grammar \cite{alegre2004probabilistic}, syntax trees \cite{ripperda2006reconstruction}, and the bottom-up or top-down hybrid approach \cite{han2008bottom}. They all required setting the correct parameters of the shape syntax in advance. Although these methods have achieved good results, they assume that the image is composed of a fairly regular grid; in addition, fixed expressions of the grammars are not capable to cover the diversities in real-world applications. Procedural grammars are also quite cumbersome to be edited and compiled, which requires tremendous expert knowledge. Human intervention is also required to select the appropriate grammar for a particular building. Although style classifiers \cite{mathias2016atlas} was developed to alleviate the above issues, which automatically recognized architectural styles from low-level image features, the use of style syntax is still needed in advance, which is probably a limitation for this approach.

Recent approaches based on mixed integer programming is arguably the most flexible and powerful tool for the problem of regular arrangement of objects. It has been used for arrangements of the 2D boundaries and 3D planes \cite{monszpart2015rapter}, reconstruction of surface meshes \cite{boulch2014piecewise,nan2017polyfit}, modeling of the roof structures of the LOD-2 models \cite{goebbels2019beautification} and the fa\c{c}ades \cite{hensel2019facade}. However, most of them formulated the optimization problem as MILP \cite{goebbels2019beautification,hensel2019facade} or even mixed integer non-linear programming \cite{monszpart2015rapter}, which has unknowns in both spaces of integer and real values. Unfortunately, this kind of problems raised up in the operational research has no efficient solvers for large scale problems, even using state-of-the-art commercial libraries \cite{gurobi2014gurobi}. A practical remedy is to reformulate the problem into BIP \cite{nan2017polyfit,kelly2017bigsur,kelly2018simplifying}, which only considers binary variables and linear energies; the regularities can still be explicitly modeled through the logical operations using the binary variables and there are relatively more efficient solvers for these simpler problems. Therefore, we use BIP to model the regularization problem of the fa\c{c}ade objects.

\section{Detection of fa\c{c}ade primitives using YOLOv3}
\label{s:yolo}

We use YOLOv3 \cite{redmon2018yolov3} to detect axis-aligned bounding boxes of primitives because of its real-time performance. For completeness, we briefly introduce the architecture and implementation details of YOLOv3 here. Rather than other region-based CNN methods \cite{girshick2015region}, YOLO \cite{redmon2016you} uses regression to directly process the entire image, and obtains categories and positions of the targets in a single forward propagation. YOLO implements an end-to-end pipeline for detection by dividing the image into $ s \times s $ grids. If the center of the semantic component is in a grid, the grid is responsible for predicting the target. Each grid will generate $ B $ bounding boxes, and each bounding box must predict its confidence $ \chi $, which is defined as the product of the probability $ P $ of the target contained in the bounding box and the accuracy $ Q $, as $ \chi=P \times Q $. If the grid contains semantic objects, then $ P=1 $, otherwise $ P=0 $. $ Q $ represents the intersection ratio of the labeled box in training samples and the predicted box. When $ Q=1 $, it means that the labeled box and the predicted box coincide perfectly.

If a grid contains semantic components, which corresponds to $C$ classes, it is represented by $ P_i $ for each class. Therefore, we can obtain the intermediate score of each grid and each class as $ \phi_i = P_i \times \chi$. The scores are truncated at $ 0 $ and non-maximum suppression is used to remove bounding boxes with a large repetition rate. In the end, each bounding box only retains the objects with positive confidence scores and the highest categories. In YOLOv3, in order to improve the accuracy of target detection, the residual network \cite{he2016deep} is used as backbone. The features before entering the residual box and the features output by the residual box are combined to extract deeper feature information. On the building fa\c{c}ade, even if they are the same type of semantic objects, their sizes and poses are not the same. YOLOv3 uses multi-scale fusion \cite{lin2017feature} to detect objects, and has good adaptability to the scale changes of objects.

\section{Regular arrangements of fa\c{c}ade primitives using binary integer programming}

After initial extraction of the bounding boxes of the building fa\c{c}ade, we then use BIP to restore the spatial regularity of the windows, doors and balconies, inspired by previous work \cite{hensel2019facade}. Although the MILP method has been successfully used in many studies \cite{boulch2014piecewise,hensel2019facade}, in our pipeline, because we are aiming at an interactive reconstruction pipeline, the runtime should be kept reasonably low. In the following, we describe our reformulated problem setup using BIP instead of MILP.

\subsection{Problem setup using binary integer programming}

\begin{figure}[h]
    \subcaptionbox{Parameters for primitives}[\columnwidth]{\includegraphics[width=0.9\columnwidth]{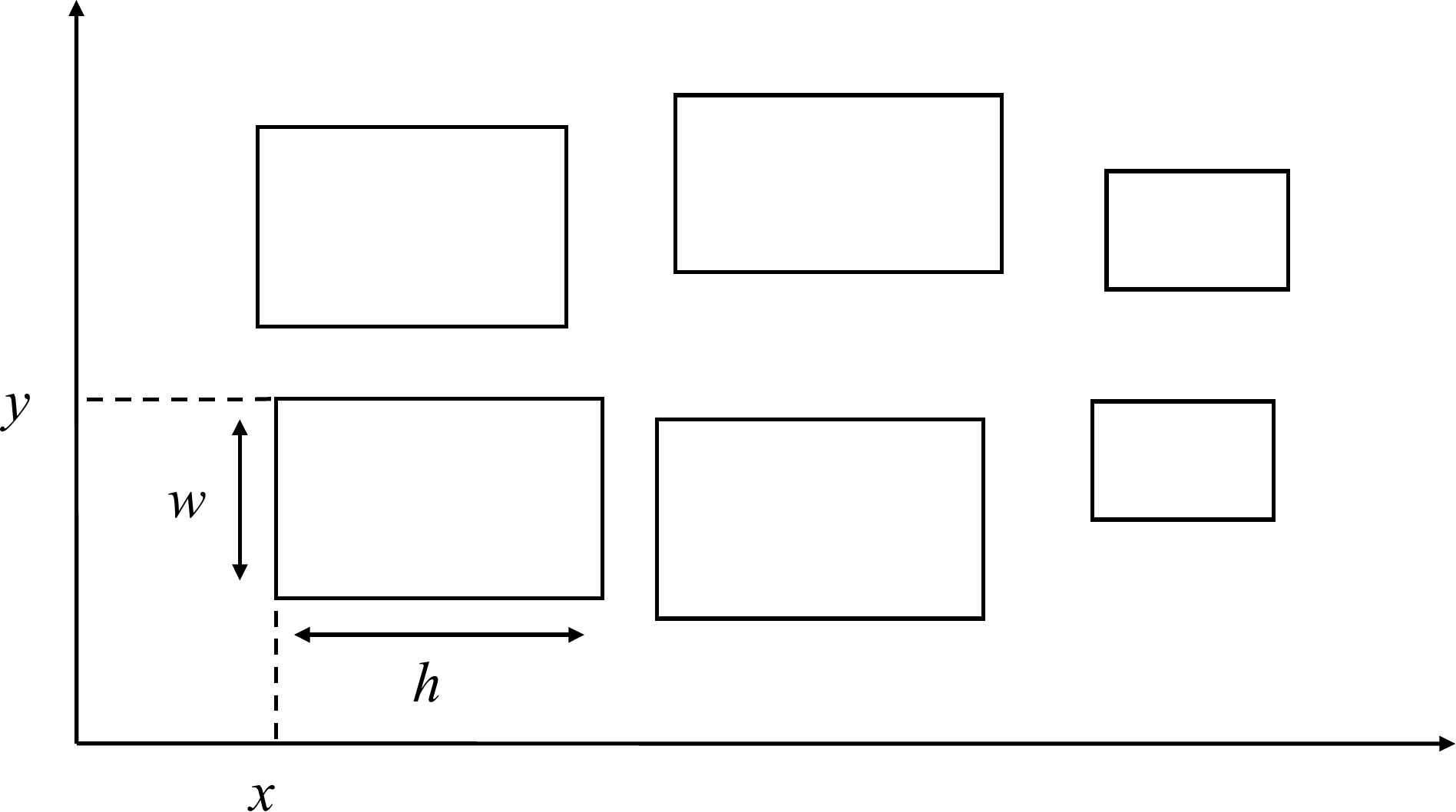}}
    \subcaptionbox{Parameters for model spaces}[\columnwidth]{\includegraphics[width=0.9\columnwidth]{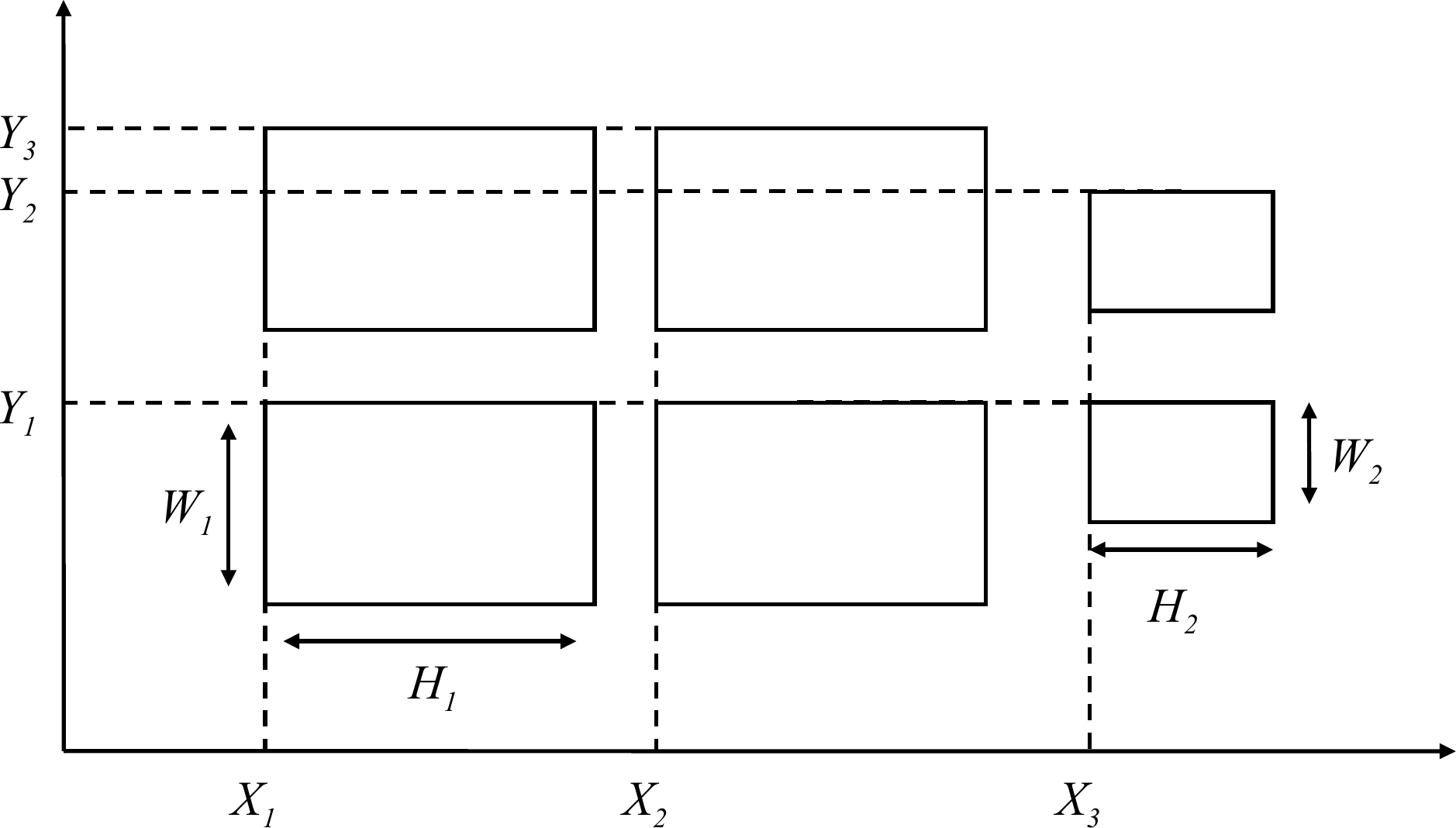}}
    \caption{Parameters for primitives and model spaces.}
    \label{fig:param}
\end{figure}

After extracting the initial primitives, we have $ N $ bounding boxes for each image, and each bounding box is uniquely determined by a set of four parameters $ (x,y,w,h) $ , where $ (x,y) $ and $(w,h)$ are coordinate of the upper left corner and size of the bounding box, respectively (Figure \ref{fig:param}a). Instead of directly optimizing these parameters that are real values using MILP \cite{hensel2019facade}, we cast it into a model selection problem using BIP.

Specifically, we first establish a model space for each \emph{attribute} of the bounding box, \emph{e.g.} $ \textbf{X} = \{ X_1, X_2, ... , X_N \} $ for the attribute of $ x $ coordinate. The size of $ |\textbf{X}| $ could be the number of bounding boxes $ N $, but we choose to compress it by pre-cluster the model space using a very confident lower bound $ \delta_l $ as described later. We then assign a binary variable $ a_{i,k}^x \in \{0,1\} $ to represent the state of the selection, \emph{i.e.} if the model $ X_k $ is selected for the attribute $ x $ of the $ i_{th} $ bounding box. In addition, we use the one-hot vector $ \boldsymbol{\xi}_i $ \footnote{We omit the superscript for attribute when not ambiguous. In addition the Greek symbols are used for one-hot vectors and Roman symbols for variables.} to represent the whole state of the $ i_{th} $ bounding box as $ \boldsymbol{\xi}_i=(a_{i,0}, a_{i,1}, ..., a_{i,|\textbf{X}|})^T$.

In fact, the model spaces of the attributes of the primitives for a single fa\c{c}ade are generally quite limited in urban environment. That is the ratio $ N/|\textbf{X}| $ is generally quite large, which leads to unnecessarily too many parameters. Therefore, we pre-cluster all the attributes separately using the mean shift approach \cite{cheng1995mean}; and the threshold is set to the lower bound $ \delta_l $. The values in the model space $ \textbf{X} $ are determined by the centers of the clusters, as shown in Figure \ref{fig:param}b. To ensure the accuracies of the results, the lower bound $ \delta_l $ in mean shift clustering should be as small as possible to avoid aggregating parameters of different properties into the same category. It should be noted that, although the same threshold $ \delta_l $ is used for all the attributes, the number of clusters $ |\textbf{X}| $, $ |\textbf{Y}| $, $ |\textbf{W}| $ and $ |\textbf{H}| $ are generally different.

In summary, the purpose is to optimize all the selecting vectors $ \boldsymbol{\xi} $, under the energy functions and constraints as described below. And the total size of explicit unknowns is $ N \times (|\textbf{X}| + |\textbf{Y}| + |\textbf{W}| + |\textbf{H}|) $.

\subsection{Energy functions to be optimized}

Our loss function consists of a data item and a regularity item. First of all, our goal is to make the sum of the changes of the bounding boxes against the initial locations as small as possible after the regularization. Therefore, we first calculate the residual vector $ \boldsymbol{\epsilon} $ for each bounding box, which represents the errors for different selections, as
\begin{equation}
    \boldsymbol{\epsilon}_i^x = (x_i - X_0, x_i - X_1, ..., x_i - X_{|X|})^T,
\end{equation}
where the superscript $ x $ denotes different attributes.

In this way, the total energy $ O_d^a $ for attribute $ a $ caused by the selection vectors, \emph{e.g.} offsets for the coordinates of upper left corners and differences for the sizes of the rectangles, can be briefly expressed as,
\begin{equation}
    O_d^a=\sum_i^N|\boldsymbol{\epsilon}^a_i| \cdot \boldsymbol{\xi}^a_i.
    \label{eq:O_d^a}
\end{equation}
Equation \ref{eq:O_d^a} means that, for each bounding box, we only account for the error of the selected value in model space, \emph{i.e.} when $ a_{i,k}=1 $. The final data term of the energy function is therefore intuitively the summation of all the attributes as 
\begin{equation}
    O_d=O_d^x+O_d^y+O_d^w+O_d^h.
    \label{eq:O_d}
\end{equation}

\begin{figure}[h]
    \centering
    \includegraphics[width=0.6\columnwidth]{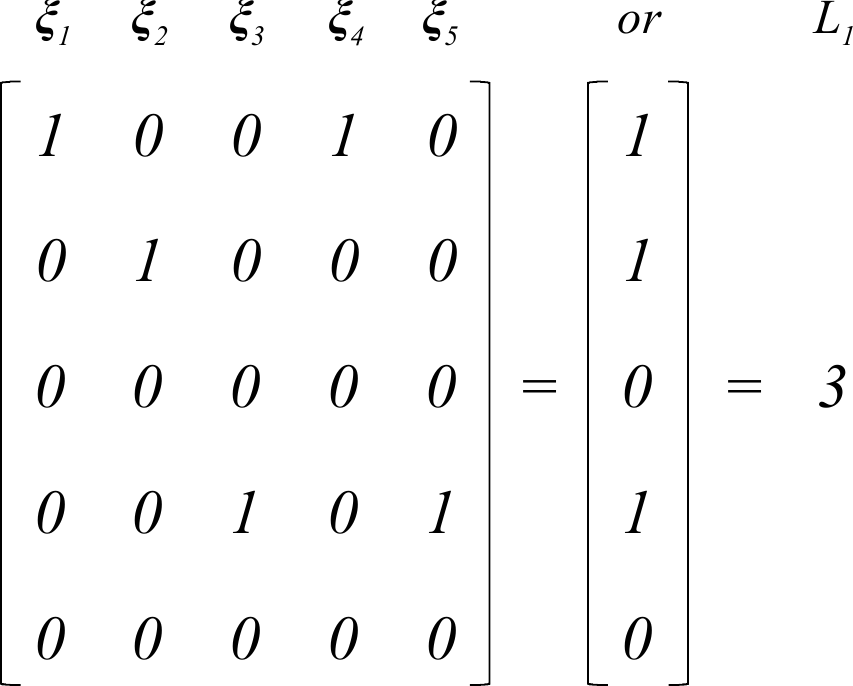}
    \caption{Illustration of the regularity energy as in Equation \ref{eq:energy_regularity}, which is the total number of selected models.}
    \label{fig:energy_regularity}
\end{figure}

With only the data term, we always have a trivial solution that have the best fit, \emph{e.g.} choosing the nearest center of the mean shift clustering. Therefore, we introduce a regularity item. The intuition behind this term is that higher regularity generally means less categories; fortunately, the number of selected categories is easy to model as illustrated in Figure \ref{fig:energy_regularity}. For each attribute $ a $, the total number of selected categories, \emph{e.g.} the regularity term $ O_g^a $, can be explicitly expressed as the following logical expression,
\begin{equation}
    O_g^a=\|\boldsymbol{\xi}_1^a \lor \boldsymbol{\xi}_2^a \lor ... \lor  \boldsymbol{\xi}_N^a\|_1,
    \label{eq:energy_regularity}
\end{equation}
where $ \| \cdot \|_1 $ is the $L_1$ norm that is the absolute summation of all the elements of a vector and for binary variables $L_1$ norm simply counts the number of non-zero variables; the binary operator $ \lor $ is the element-wise \emph{logical or} for the one-hot vectors. Similar to Equation \ref{eq:O_d}, the final regularity term is a weighted summation across all the attributes as 
\begin{equation}
    O_g=\omega^x O_g^x+\omega^y O_g^y+\omega^w O_g^w+\omega^h O_g^h,
\end{equation} 
where $\omega$ denotes the weights of different attributes. And the final energy function is 
\begin{equation}
    O=O_d+O_g.
\end{equation} 

\subsection{Constraints of the binary integer programming}

The variables $ \boldsymbol{\xi}_i=(a_{i,0}, a_{i,1}, ..., a_{i,N})^T $ cannot be adjusted freely. Obviously, because each bounding box can only choose one state, we have the following constraint $ C_1 $ for each bounding box,
\begin{equation}
    C_1:\sum_k{a_{i,k}}=1,\forall ~ i \leq N.
\end{equation}

Another practical constraint $ c_2 $ is that we could very confidently ignore certain model spaces if the residual $ |\boldsymbol{\epsilon}_{i,k}| $ exceeds an upper bound $ \delta_u $, as.
\begin{equation}
    C_2:a_{i,k}=0,~ \forall ~ |\boldsymbol{\epsilon}_{i,k}|>\delta_u.
\end{equation}
It seems the additional constraints may increase the complexity of the problem, but interestingly, in practice, we find that the additional constraints significantly reduce the runtime, with almost no differences in the final results. 

\subsection{Implementation details}

The implementation of Equation \ref{eq:energy_regularity} needs some tricks, because it involves the logical operations. For two binary variable $ a $ and $ b $, the \emph{logical or} result $ c=a \lor b $ can be modeled by adding the following constraints,
\begin{equation}
    \begin{split}
        c \leq& a+b \\
        c \geq& a \\
        c \geq& b
    \end{split}.
\end{equation}
In fact, this kind of fixed routines can be handled efficiently and gracefully by state-of-the-art solvers \cite{chinneck2007feasibility,gurobi2014gurobi}. For the parameters, we set $\delta_l\in[3,5]$ pixels and $ \delta_u=10\delta_l$; and $ \omega^x=\omega^y=100 $ and $ \omega^w=\omega^h=10 $ are used empirically. In this way, all the energy functions and constraints are linear functions, which are solved using the Mosek library \cite{mosek2010mosek}.

\section{Experimental evaluations}

\subsection{Evaluation of detections of fa\c{c}ade primitives}

This paper uses the CMP fa\c{c}ade database \cite{tylecek2012cmp} as the training data set, which contains a total of 606 building fa\c{c}ade images around the world. These images are manually labeled with 12 semantic objects on the fa\c{c}ade. We choose three typical primitives: window, door and balcony. We built the YOLOv3 model based on Keras \cite{gulli2017deep} to train the above data set. At the same time, we took 30 typical building fa\c{c}ade images from Google street view \cite{anguelov2010google} for testing, and manually labeled them for evaluations. In order to verify the effectiveness of this method, we adopted the same evaluation method in \cite{rahmani2018high}. We counted every classified pixel as either true positive(TP) or false positive(FP), and the precision is thus $TP/(TP+FP)$. On our test dataset, for windows, doors, and balconies, our average extraction precision reached 0.917, 0.856, and 0.852. 

In addition, we used the same test dataset ICG Graz50 \cite{riemenschneider2012irregular} to compare with a recent method \cite{hensel2019facade} based on Fatser RCNN (Table \ref{tab:detection}). Both methods are trained on the CMP dataset. The precisions of the extraction of windows and doors listed in \cite{hensel2019facade} are 0.892 and 0.834, and the precision of our method based on YOLOv3 are 0.882 and 0.825. Considering that YOLOv3 is more efficient than Faster RCNN, the mild precision loss is acceptable. And the detection performance could be considered satisfactory.

\begin{table}[h]
    \centering
    \caption{Comparison of precisions on the ICG Graz50 dataset, with model trained on the CMP fa\c{c}ade dataset.}
    \label{tab:detection}
    \begin{tabular}{@{}lll@{}}
        \toprule
        & Window & Door  \\ \midrule
        Hensel et al. (2019) & 0.892  & 0.834 \\
        Proposed             & 0.882  & 0.825 \\ \bottomrule
    \end{tabular}%
\end{table}

We tested the precision of window extraction using images with different resolutions and different layout complexity (Table \ref{tab:scale-complexity}). The scale is measured by downsampling the images and the layout complexity is the total number of the selected modes on the of the four attributes of the windows. It can be seen from Table \ref{tab:scale-complexity} that when the resolution of the image is within a certain range, the extraction precision is good and the difference is small; but when the resolution is too low, the extraction precision is significantly reduced. In addition, it can be noticed that as the layout complexity increases, the extraction precision tends to gradually decrease. In summary, when the images are captured at a relatively high resolution, the layout complexity has a greater impact on the extraction precision.

\begin{table}[h]
    \centering
    \caption{Detection performance with respect to different scale of images and different complexities of fa\c{c}ade layout. }
    \label{tab:scale-complexity}
     \begin{tabular}{@{}lccccc@{}}
        \toprule
        Scale      & 1     & 1/2   & 1/4   & 1/8   & 1/16  \\
        Precision  & 0.854 & 0.845 & 0.893 & 0.895 & 0.565 \\ \midrule
        Complexity & 32    & 36    & 56    & 76    & 92    \\
        Precision  & 0.928 & 0.916 & 0.908 & 0.868 & 0.82  \\ \bottomrule
    \end{tabular}%
\end{table}

\vspace{1em}

\subsection{Evaluation and comparisons of the regular arrangements of the primitives}

We selected three typical building fa\c{c}ade images of three cities in the United States (US), United Kingdom (UK), and Canada (CA) to evaluate the performance of the regularization. Both qualitative and quantitative evaluations are conducted and we also compare the runtime performance against the MILP approach \cite{hensel2019facade}.

\begin{figure*}[htbp]
	\centering
	\subcaptionbox{CA}[0.32\linewidth]{\includegraphics[height=24cm]{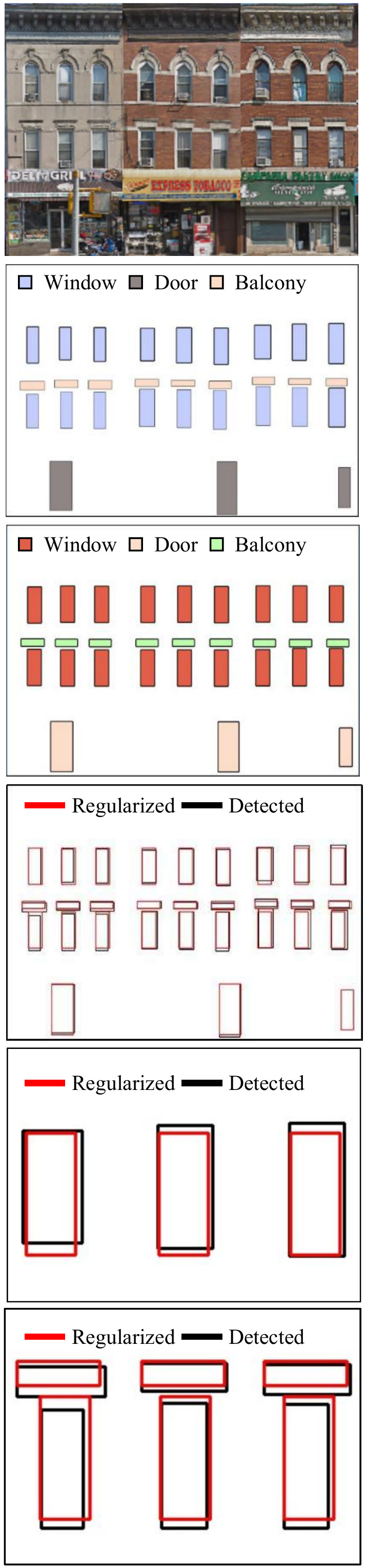}}   \subcaptionbox{UK}[0.32\linewidth]{\includegraphics[height=24cm]{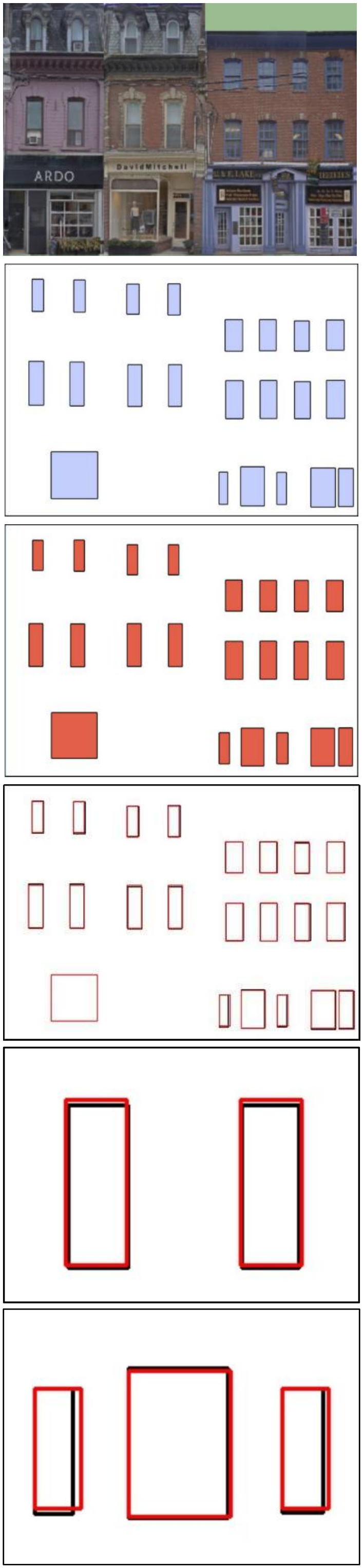}}
	\subcaptionbox{US}[0.32\linewidth]{\includegraphics[height=24cm]{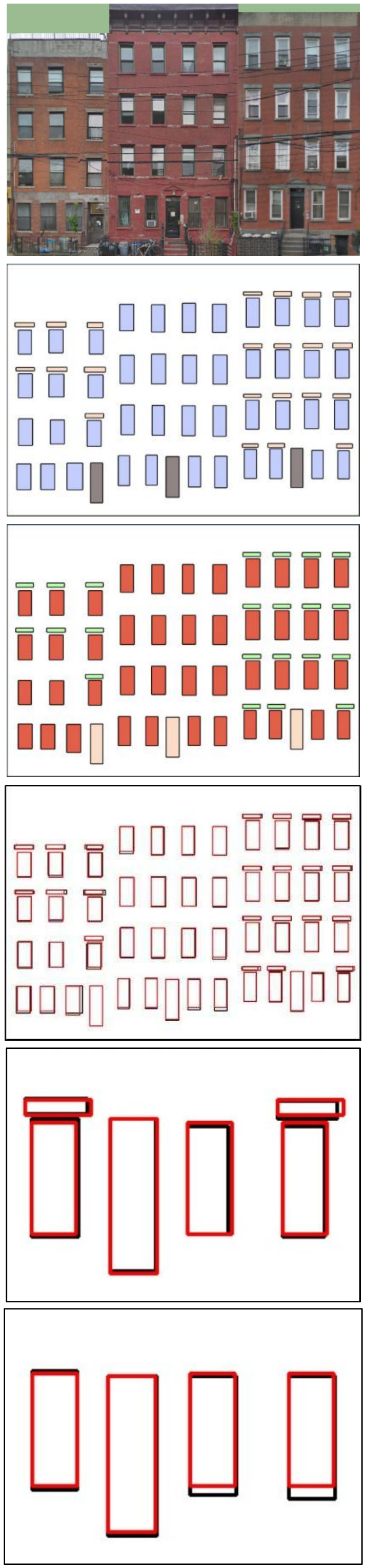}}
	\caption{Comparison of detected and regularized fa\c{c}ade primitives for the three datasets. The second and third rows show the detected and regularized primitives, respectively. The fourth row compares them and the last two rows give two enlarged demonstrations.}
	\label{fig:regularized}
\end{figure*}

\begin{figure}[htb]
    \centering
    \subcaptionbox{US}[\columnwidth]{\includegraphics[width=0.45\columnwidth]{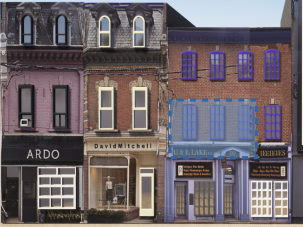} \includegraphics[width=0.45\columnwidth]{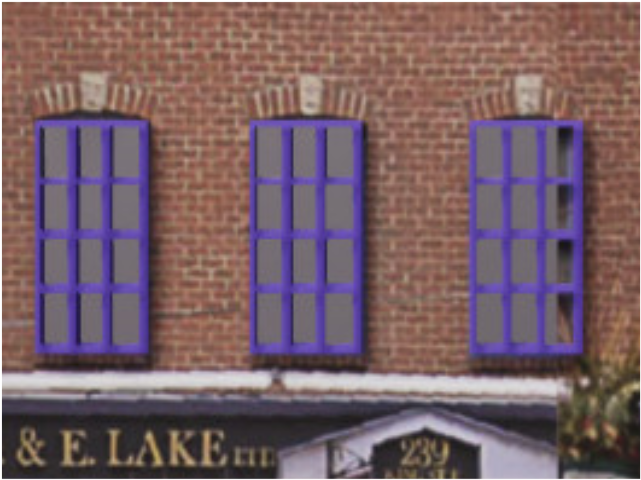}}
    \subcaptionbox{UK}[\columnwidth]{\includegraphics[width=0.45\columnwidth]{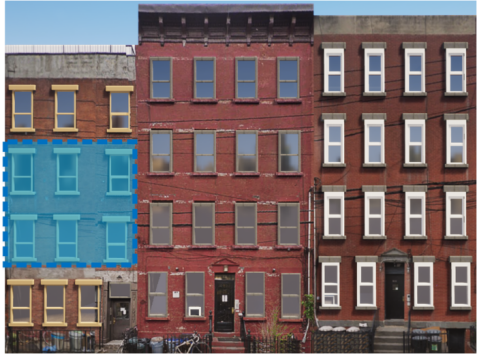} \includegraphics[width=0.45\columnwidth]{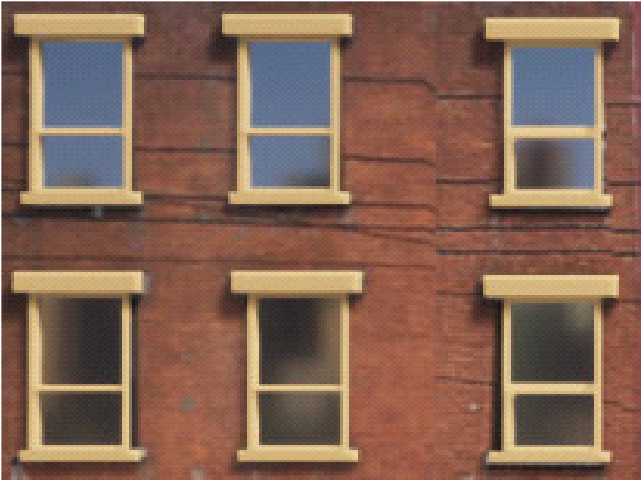}}
    \subcaptionbox{CA}[\columnwidth]{\includegraphics[width=0.45\columnwidth]{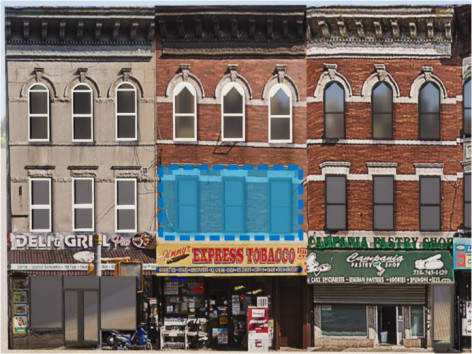} \includegraphics[width=0.45\columnwidth]{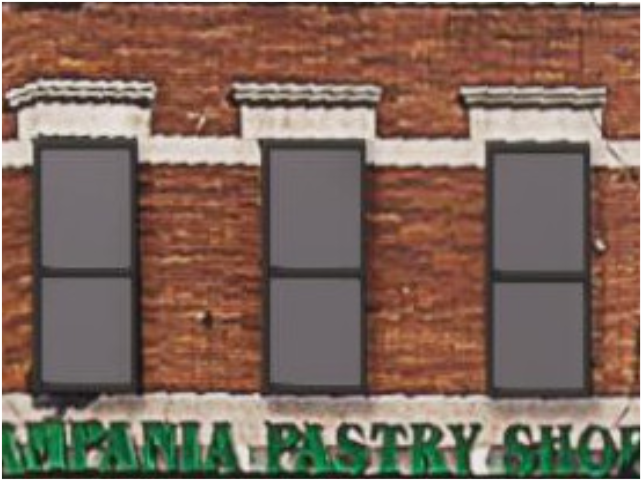}}
    \caption{Reconstructed fa\c{c}ades for the three datasets, in which the right column demonstrates the enlarged areas in the cyan rectangles.}
    \label{fig:reconstruction}
\end{figure}

\paragraph{Qualitative evaluations.} Figure \ref{fig:regularized} compares the extracted and regularized bounding boxes for the US, UK and CA datasets. The black frame represents the extracted primitives and the red frame indicates the regularized results. It can be noticed that after regularization, the semantic objects on the building fa\c{c}ade are arranged more neatly and consistently and still fit well enough to the original bounding boxes, as demonstrated in the enlarged regions. In addition, Figure \ref{fig:reconstruction} demonstrates the reconstructed fa\c{c}ades for the three datasets in off-the-shelf modeling solutions.

\paragraph{Quantitative evaluations.}

We counted the number of used model space before and after the regularization to measure the regularity of the results, e.g. $O_g^a$ in Eq. \ref{eq:energy_regularity}. Table \ref{tab:num_params} demonstrates the results, and it could be noted that, the selected parameters only account for about $ 50\% $ for the coordinates of the corners and $ 20\% $ for the sizes.

\begin{table}[h]
    \centering
    \caption{Quantitative evaluations of the regularity by the size of the model space before and after regularization, \emph{i.e.} $O_g^a$ in Eq. \ref{eq:energy_regularity}. The number of the selected model space significantly reduced after optimization.}
    \label{tab:num_params}
    \begin{tabular}{@{}ccccccccc@{}}
    \toprule
    \multirow{2}{*}{Dataset} & \multicolumn{4}{c}{\#Detected} & \multicolumn{4}{c}{\#Regularized} \\
                             & $|\textbf{X}|$           & $|\textbf{Y}|$          & $|\textbf{W}|$          & $|\textbf{H}|$          & $|\textbf{X}|$           & $|\textbf{Y}|$          & $|\textbf{W}|$          & $|\textbf{H}|$           \\ \midrule
    US                       & 76          & 62         & 35         & 39         & 38           & 25           & 5            & 4           \\
    UK                       & 22          & 20         & 17         & 17         & 16           & 6            & 6            & 6           \\
    CA                       & 47          & 39         & 29         & 26         & 31           & 6            & 10           & 5           \\ \bottomrule
    \end{tabular}
\end{table}

\paragraph{Comparisons of runtime.}

In order to verify the efficiency of the method in this paper, we tested six building fa\c{c}ades with complex structures and numerous parameters, and compared the proposed BIP approach against the MILP approach \cite{hensel2019facade}. The results are shown in the Table \ref{tab:runtime} and the runtime of the proposed BIP approach only accounts for about $10\%$ to $ 20\% $ of the MILP approach. For the MILP approach \cite{hensel2019facade}, the number of explicit unknown parameters are $ N(2|\textbf{X}|+2|\textbf{Y}|+|\textbf{W}|+|\textbf{H}|) + 8N $, including $ 8N $ real value parameters. In the proposed approach, the number of explicit unknown parameters is $ N(|\textbf{X}|+|\textbf{Y}|+|\textbf{W}|+|\textbf{H}|) $. Although the proposed method has slightly fewer parameters, the numbers are still in the same order of magnitude. Therefore, it is the reformulated problem that account for the performance differences.

\begin{table}[h]
    \centering
    \caption{Comparison of the runtime between MILP and the proposed BIP approaches. The second to fifth columns demonstrates the complexities of the model space.}
    \label{tab:runtime}
    \begin{tabular}{@{}ccccccc@{}}
    \toprule
    $N$  & $|\textbf{X}|$           & $|\textbf{Y}|$          & $|\textbf{W}|$          & $|\textbf{H}|$           & MILP (s)  & BIP (s)  \\ \midrule
    26 & 11 & 5  & 3 & 2 & 5.7   & 0.9  \\
    74 & 20 & 13 & 3 & 3 & 150.9 & 19.9 \\
    60 & 29 & 10 & 4 & 7 & 135.2 & 20.8 \\
    61 & 10 & 16 & 4 & 7 & 84.6  & 12.7 \\
    67 & 24 & 6  & 4 & 5 & 106.2 & 16.6 \\
    45 & 35 & 12 & 9 & 9 & 123.6 & 20.3 \\ \bottomrule
    \end{tabular}
\end{table}

\section{Conclusion}

This paper proposed an approach for the regular arrangement of primitives of the building fa\c{c}ades using BIP. Compared to the MILP approach, BIP is considerably faster and achieves near real-time performance with similar level of data fitness and regularities. The detected and rearranged bounding boxes of the primitives can be directly used for the modeling of the fa\c{c}ade features, which is a key step towards the LOD-3 reconstruction. However, current approaches can only detect axis-aligned objects, future works may be devoted to explore the reconstruction of more complex fa\c{c}ade features. Code is available at https://github.com/saedrna/Ranger.

\section*{Ackowledgement}

This paper is supported by the National Natural Science Foundation of China (Project No.: 41631174, 61602392, 41871291).

{
	\begin{spacing}{1.17}
		\normalsize
        \bibliography{RegularFacade}
	\end{spacing}
}

\end{document}